\documentclass[letterpaper, 10 pt, conference]{ieeeconf}  

\usepackage{cite}
\usepackage{amsmath,amssymb,amsfonts}
\usepackage{float}
\usepackage{cuted}
\usepackage{capt-of}
\usepackage{algorithmic}
\usepackage{graphicx}
\usepackage{textcomp}
\usepackage{xcolor}
\usepackage{colortbl}
\usepackage{algorithm}
\usepackage{algorithmic}
\usepackage{booktabs} 
\usepackage{multirow}
\usepackage{pifont}
\newcommand{\cmark}{\ding{51}}
\newcommand{\xmark}{\ding{55}}
\def\BibTeX{{\rm B\kern-.05em{\sc i\kern-.025em b}\kern-.08em
    T\kern-.1667em\lower.7ex\hbox{E}\kern-.125emX}}
\begin{document}

\title{Closing the Navigation Compliance Gap in End-to-end Autonomous Driving\\
}
\makeatletter
\newcommand{\linebreakand}{%
  \end{@IEEEauthorhalign}
  \hfill\mbox{}\par
  \mbox{}\hfill\begin{@IEEEauthorhalign}
}
\makeatother
\author{
\authorblockN{1\textsuperscript{st} Hanfeng Wu}
\authorblockA{\textit{Karlsruhe Institute of Technology} \\
\textit{BMW Group}\\
Munich, Germany \\
hanfeng.wu@bmw.de}
\and
\authorblockN{2\textsuperscript{nd} Marlon Steiner}
\authorblockA{\textit{Karlsruhe Institute of Technology} \\
Karlsruhe, Germany \\
marlon.steiner@kit.edu}
\and
\authorblockN{3\textsuperscript{th} Michael Schmidt}
\authorblockA{\textit{BMW Group} \\
Munich, Germany \\
michael.se.schmidt@bmw.de}
\linebreakand
\authorblockN{4\textsuperscript{rd} Alvaro Marcos-Ramiro$^1$}
\authorblockA{\textit{BMW Group} \\
Munich, Germany \\
alvaro.marcos-ramiro@bmw.de}
\and
\authorblockN{5\textsuperscript{th} Christoph Stiller}
\authorblockA{\textit{Karlsruhe Institute of Technology} \\
Karlsruhe, Germany \\
stiller@kit.edu}
}

\twocolumn[{%
\renewcommand\twocolumn[1][]{#1}%
\maketitle
\vspace{-6mm}
\begin{center}
    \centering
    \includegraphics[width=\textwidth]{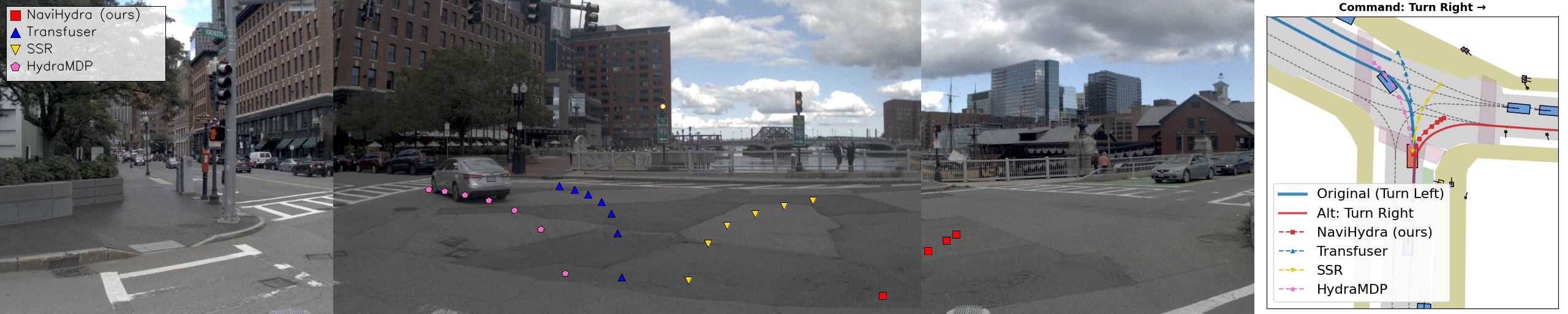}
    \vspace{-6mm}
    \captionof{figure}{Illustration of model predictions with alternative navigation command. The ego vehicle, shown in red, is at the entrance of an intersection. The original route is \textbf{turning left}. We demonstrate the models' predictions under an alternative command~(\textbf{turning right}). Our method responds correctly to the driving command, while Hydra-MDP and Transfuser following the original driving route, SSR deviating from the driving path.}
    \label{fig:teaser}
\end{center}%
}]
\footnotetext[1]{denotes corresponding author}
\begin{abstract}

Trajectory-scoring planners achieve high navigation compliance when following the expert's original command, yet they struggle at intersections when presented with alternative commands—over 30\% of such commands are ignored. We attribute this \emph{navigation compliance gap} to two root causes: (1)~existing metrics like Ego Progress do not explicitly measure navigation adherence, diluting the gap between on-route and off-route trajectories; and (2)~current datasets pair each scenario with a single command, preventing models from learning command-dependent behavior.
We address the metric gap by introducing the binary Navigation Compliance metric~(NAVI) and the derived Controllability Measure~(CM), and the data gap with the \emph{NavControl} dataset—14,918 intersection scenarios augmented with all feasible alternative commands and routing annotations, yielding over 34,000 direction samples.
Building on these, we propose NaviHydra, a trajectory-scoring planner incorporating NAVI distillation and Bird's Eye View~(BEV)-based trajectory gathering for context-position-aware trajectory feature extraction. NaviHydra achieves 92.7 PDM score on NAVSIM navtest split and 77.5 CM on \textbf{NavControl} test split. Training with \textbf{NavControl} improves controllability across diverse architectures, confirming it as a broadly effective augmentation for navigation compliance. Our project page is https://hanfeng-wu.github.io/closing/.
\end{abstract}

\section{Introduction}
Consider an autonomous vehicle approaching an intersection, as shown in Fig.~\ref{fig:teaser}, where the driver's navigation system indicates to turn right. A trajectory-scoring planner evaluates thousands of candidate trajectories and selects the one with the highest safety score—but that trajectory turns left, ignoring the navigation command entirely. This is not a hypothetical scenario: when we evaluate the state-of-the-art trajectory-scoring method Hydra-MDP~\cite{li2024hydra,li2025generalized} on intersection scenarios with extra alternative navigation commands, more than 30\% of these driving commands are not followed—the planner selects a trajectory that ignores the commanded direction entirely. We term this the \textbf{navigation compliance gap}: existing closed-loop metrics reward safe forward motion but fail to effectively penalize trajectories that violate navigation intent.

The root cause is twofold. First, the metrics used to supervise trajectory selection provide insufficient navigation supervision. Traditional end-to-end (E2E) planners~\cite{hu2023planning,chitta2022transfuser} employ imitation learning, which implicitly couples navigation intent with expert demonstrations but suffers from unsafe interpolation~\cite{chen2024vadv2}. Trajectory-scoring methods~\cite{li2024hydra,li2025hydramdpadvancingendtoenddriving,li2025generalized} improve safety by scoring a fixed set of proposals using closed-loop simulation metrics (collision avoidance, drivable area compliance, comfort, etc.). Among these, Ego Progress (EP) is the metric most related to navigation: it measures how far the trajectory's endpoint projects onto the expert's centerline, which implicitly rewards trajectories that stay near the expert's path. However, EP falls short as a navigation compliance signal: it measures geometric proximity to the nearest centerline point rather than verifying that the trajectory reaches the intended destination, and its normalization into a relative ranking with partial credit further dilutes the gap between on-route and off-route trajectories. Without a dedicated binary metric for route adherence, the model lacks a clear gradient signal to learn navigation-compliant behavior.

\begin{figure}
    \centering
    \includegraphics[width=1\linewidth]{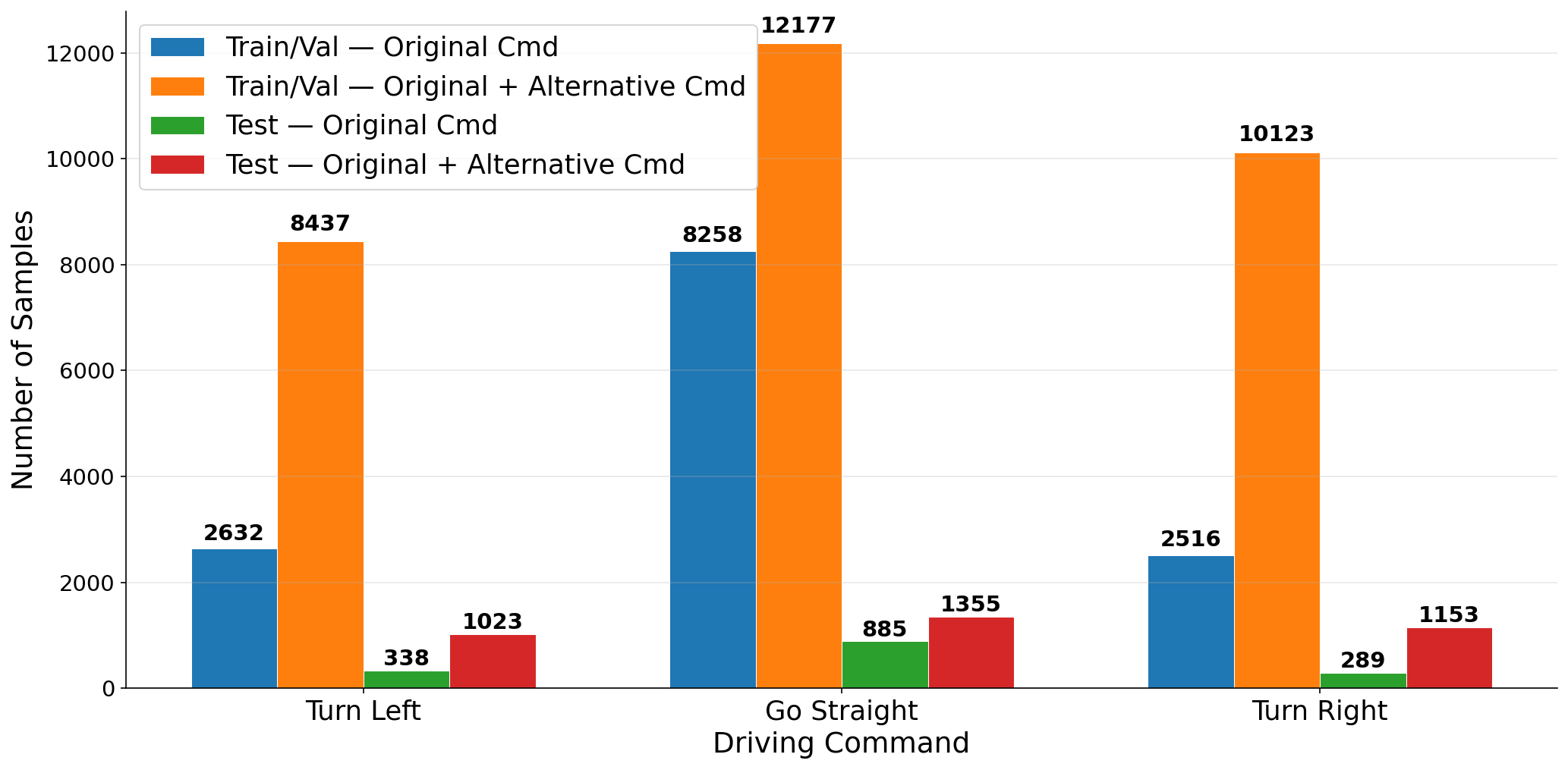}
    \vspace{-6mm}
    \caption{Distribution of driving commands in \textbf{NavControl} dataset for training/validation and testing. We present the number of both original driving commands and alternative driving commands. The augmentation provides a more balanced distribution of driving commands, where the original distribution is dominantly straight driving.}
    \label{fig:navcontrol}
\end{figure}

Second, existing training datasets and benchmarks lack the \emph{intersection-level diversity} needed to teach or evaluate command-following. In the standard NAVSIM splits, each scenario is paired with a single navigation command—the one the expert actually executed. The model never observes the same intersection under alternative commands (e.g., ``turn left'' vs.\ ``go straight'' vs.\ ``turn right''), so it cannot learn to differentiate its behavior accordingly. Evaluation suffers from the same limitation: since each scenario is tested under only one command, there is no mechanism to verify whether the planner's output would change had a different command been issued.

To close both gaps simultaneously, we construct the \textbf{NavControl} dataset, a purpose-built augmentation of the NAVSIM benchmark that provides 14,918 intersection scenarios, each annotated with all geometrically feasible alternative navigation commands and their corresponding routing lanes. 
The \textbf{NavControl} dataset serves a dual purpose: as \emph{training data}, it exposes the model to the same scene under different commands, providing the contrastive signal needed to learn command-dependent behavior; and as a \emph{test benchmark}, it enables rigorous evaluation of controllability by presenting each scene with multiple commands and checking whether the planner's output changes accordingly.

Building on this dataset, we propose the \textbf{Controllability Measure (CM)}, a benchmark protocol that quantifies a model's responsiveness to different navigation commands at the same scene. CM leverages the \textbf{Navigation Compliance metric (NAVI)}, which verifies whether a trajectory's last waypoint lies on the commanded route—and multiplies it with the PDM score, so that only trajectories that are both navigation-compliant \emph{and} safe receive credit. This makes the CM a strong indicator of whether a planner truly follows commands or merely imitates dominant driving patterns.

To fully exploit the \textbf{NavControl} dataset, we introduce \textbf{NaviHydra}, a navigation-guided trajectory-scoring framework. NaviHydra incorporates the NAVI metric as a distillation head, treating route adherence as a first-class closed-loop metric on par with safety and comfort. We further propose \textbf{trajectory gathering}, a BEV-based mechanism that constructs trajectory features by sampling spatial representations along each proposal's waypoints, providing richer context for more informed scoring decisions.

Extensive experiments demonstrate that training with the augmented intersection data from \textbf{NavControl} consistently improves the Controllability Measure across many evaluated methods—spanning both trajectory-scoring approaches and imitation-learning-based methods—confirming that the \textbf{NavControl} dataset is a broadly effective augmentation that enables end-to-end models of diverse architectures to react appropriately to different driving commands. Furthermore, NaviHydra achieves state-of-the-art performance with a 92.7 PDM score on navtest and 77.5 CM on \textbf{NavControl}. In summary, our contributions are:


\begin{itemize}
    \item We identify the \textbf{navigation compliance gap} in trajectory-scoring planners, caused jointly by the absence of an explicit route-adherence metric and the lack of intersection-level diversity in existing datasets.
    \item We construct the \textbf{NavControl} dataset, augmenting 14,918 intersection scenarios with permissible alternative commands and routing annotations, serving as both a training augmentation and a test benchmark.
    \item We introduce the \textbf{NAVI} metric (Sec.~\ref{sec:navi}) and the derived \textbf{Controllability Measure (CM)} $=NAVI \times PDM$, rewarding only trajectories that both follow the commanded direction and drive safely.
    \item We propose \textbf{trajectory gathering} (Sec.~\ref{sec:td}) which, together with NAVI, extends Hydra-MDP\cite{li2024hydra} to the \textbf{NaviHydra} framework, achieving state-of-the-art performance on both the NAVSIM and the \textbf{NavControl} benchmark.
\end{itemize}

\section{Related work}
\subsection{Benchmarks and metrics for autonomous driving}
Progress in autonomous driving heavily depends on high‑quality evaluation benchmarks. nuScenes~\cite{caesar2020nuscenes} provides multimodal 3D annotations for perception but lacks a planning benchmark. nuPlan~\cite{nuplan} offers closed-loop planning evaluation on real data but is limited by the computational cost of reactive rollouts. The CARLA simulator~\cite{dosovitskiy2017carla} evaluates goal-directed navigation via route completion and infractions but tests only a single fixed route per episode. WOD-E2E~\cite{xu2025wode2e} curates long-tail driving segments and proposes the human-aligned Rater Feedback Score~(RFS), but requires costly per-scenario rater annotations and evaluates each segment under a single routing command. NAVSIM~\cite{Dauner2024NEURIPS} proposes an efficient non-reactive evaluation framework with the PDM score, decomposing driving performance into safety (NC, DAC, TTC), progress (EP), and comfort (C) sub-scores. NAVSIMv2~\cite{cao2025navsimv2} extends this with pseudo-simulation via 3D Gaussian Splatting~\cite{kerbl3Dgaussians}, achieving higher correlation with reactive simulation while retaining open-loop efficiency.

Despite these advances, none of the above benchmarks explicitly evaluates \emph{navigation controllability}—the ability of a planner to respond to different high-level commands at the same scene. All evaluation splits present each scenario under a single navigation command or goal point, making it difficult to distinguish whether a planner has learned to follow navigation commands or has merely learned a mapping from the current scene context to the ground-truth trajectory, disregarding the command entirely. Our \textbf{NavControl} dataset and the associated NAVI/CM metrics are designed to fill precisely this gap.
\vspace{-2mm}
\subsection{End-to-end autonomous driving}
End-to-end autonomous driving~(E2EAD) methods~\cite{hu2023planning,weng2024drive,winter2025bevdriver,linavigation, RAD, xing2025goalflow} map raw sensor data directly to planning outputs. BEV-based~\cite{lssinst,huang2021bevdet,liu2022bevfusion,li2024bevformer,qin2023unifusion} methods build dense spatial representations from multi-camera input, while sparse approaches~\cite{li2024hydra,li2025hydramdpadvancingendtoenddriving,chen2024vadv2,li2024enhancing,hamdan2024carformer,li2025hydranext} offer competitive performance at lower cost. On the planning side, early work~\cite{hu2023planning,jiang2023vad,chitta2022transfuser} relied on L2 imitation of expert trajectories, which can produce dangerous interpolated trajectories~\cite{chen2024vadv2}. Trajectory-scoring frameworks such as Hydra-MDP~\cite{li2024hydra,li2025hydranext} address this by distilling closed-loop simulation results into a learned scorer over trajectory proposals, replacing direct regression with classification-based training.
\vspace{-2mm}
\subsection{Navigation-guided autonomous driving}
Navigation commands serve as critical input signals for autonomous driving functions. Methods with privileged perception inputs, such as IDM~\cite{PhysRevE.62.1805} and PDM-closed~\cite{dauner2023parting}, leverage the route information directly to navigate the ego vehicle. Previous work~\cite{li2024hydra,li2023ego,chitta2022transfuser} primarily encoded navigation commands as embeddings, supplementing other ego status features. However, this simplistic integration lacks the controllability necessary to respond to different navigation commands. SSR~\cite{linavigation} successfully incorporates navigation commands into the BEV embedding using a Squeeze-and-Excitation (SE) Layer~\cite{DBLP:journals/corr/abs-1709-01507} and supervises the planner through imitation learning. While this approach effectively visualizes attention variations in response to different navigation commands, it falls short of demonstrating the model's navigation controllability due to its reliance on a straightforward imitation learning strategy.

\begin{figure}
    \centering
    \includegraphics[width=1\linewidth]{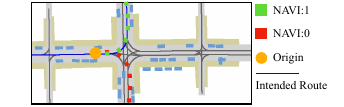}
    \vspace{-6mm}
    \caption{Illustration of Navigation Compliance metric~(NAVI): the blue lines are the lane sets on route, two evaluated trajectories colored with red and green are depicted. The green path ends on route hence has the NAVI of 1, while the red path ends off route hence it has NAVI of 0.}
    \label{fig:route}
\end{figure}

\section{NavControl dataset}\label{sec:NavControl}
As illustrated in Fig.~\ref{fig:teaser}, the \textbf{NavControl} dataset augments the NAVSIM benchmark with multi-command intersection scenarios for both training and evaluation. Its construction proceeds in two stages, followed by two associated metrics—NAVI and CM—that provide the evaluation protocol needed to close the navigation compliance gap.

\subsection{Intersection scenario selection}
Starting from every scenario in NAVSIM, we traverse the HD-map roadblock graph forward from the ego vehicle's current roadblock, inspecting up to $K_{rb}=10$ consecutive roadblocks along the planned route. At each roadblock, we enumerate all outgoing lane connectors and classify their turn direction by comparing the connector's exit heading $\theta_{out}$ with the reference heading $\theta_{ref}$ of the current route lane:
\begin{equation}
  dir(\theta_{ref},\theta_{out}) =
  \begin{cases}
    \textit{left}     & \text{if } \Delta\theta > \pi/6, \\
    \textit{right}    & \text{if } \Delta\theta < -\pi/6, \\
    \textit{straight} & \text{otherwise},
  \end{cases}
  \label{eq:dir}
\end{equation}
where $\Delta\theta = \operatorname{atan2}(\sin(\theta_{out} - \theta_{ref}),\;\cos(\theta_{out} - \theta_{ref}))$ is the signed angular difference wrapped to $[-\pi,\pi]$. The threshold $\pi/6$ (30°) symmetrically trisects the forward-facing arc $[-\pi/2,\pi/2]$ and empirically separates gentle road curvature from real intersection turns. A roadblock qualifies as an intersection when it admits at least two distinct directions. To ensure the intersection is actionable within the 4\,s evaluation horizon, we estimate the travel time from the ego position using $t = d/\max(v_{ego}, v_{min})$ with $v_{min}=5$\,m/s and discard candidates with $t > t_{max}=2$\,s. The first qualifying intersection along the route is selected. This procedure yields 14,918 qualifying scenarios comprising 34,268 direction samples in total, which we split into 13,406 scenarios (30,737 direction samples) for training/validation and 1,512 scenarios (3,531 direction samples) for testing. The concrete driving commands distribution is shown in Fig.~\ref{fig:navcontrol}.

\subsection{Alternative route construction and labeling}
For each qualifying scenario, we construct alternative routes for every feasible driving direction $c\in\{Left, Straight, Right\}$ collected from the last stage. Beginning at the identified intersection roadblock, we follow the outgoing lane connector corresponding to the alternative direction and greedily extend the route through subsequent roadblocks, accumulating lane-baseline lengths until the total distance from the ego exceeds 80 meters (roughly 4\,s at urban speed). A Dijkstra search over the resulting lane graph then produces a smooth centerline for each alternative route. 

Crucially, when scoring trajectories under an alternative command $c$, both the reference centerline and the on-route lane set are replaced with the alternative route's centerline and lane set $\mathbf{L}_{route}^c$, so that route-dependent sub-scores such as EP and NAVI are evaluated with respect to the commanded direction rather than the original route. The NAVSIM offline simulator is then re-run on all $N_t$ trajectory proposals under this substituted context, yielding per-trajectory sub-scores $\{\mathbf{S}_i^m \mid m \in M\}_{i=1}^{N_t}$ for each command, where $M = \{NC, DAC, TTC, EP, C, NAVI\}$. The resulting dataset associates every intersection scenario with multiple command--label pairs, providing the contrastive supervision signal absent from the original NAVSIM dataset.
\begin{figure*}[t]
    \centering
        \includegraphics[width=1.0\textwidth]{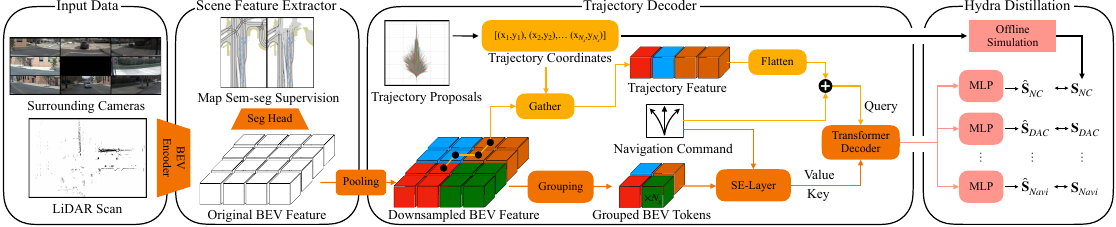}
        \vspace{-6mm}
        \caption{
        Overview of NaviHydra framework. We encode the surrounding camera images with optionally LiDAR point clouds into a BEV feature. The BEV encoder is pretrained using a semantic segmentation task. For the trajectory decoder, we gather the corresponding grid from the BEV feature using the trajectory proposal to construct the trajectory query. The navigation command is integrated into the BEV feature to create the key and value inputs. The output of the trajectory decoder is then fed into a hydra scorer to align with each sub-score from the offline simulation.
    }
    \label{fig:main}
\end{figure*}
\vspace{-4mm}
\subsection{Navigation Compliance (NAVI)}\label{sec:navi}
A key component of the \textbf{NavControl} dataset is the Navigation Compliance metric (NAVI), which we introduce to explicitly measure whether a trajectory follows the intended route. While Ego Progress (EP) implicitly relates to the expert's path by projecting trajectories onto the centerline, it yields a continuous, normalized score that gives partial credit even to off-route trajectories. NAVI instead provides a binary, absolute answer: using the stored route lane set $\mathbf{L}_{route}$, it checks whether the last waypoint of each trajectory lies on a route lane, as illustrated in Fig.
~\ref{fig:route}:
\begin{equation}
    \scalebox{0.9}{$NAVI(\mathbf{T} = \{(x^s,y^s,\theta^s)\}^{N_s}_{s=1}) = \mathbf{1}_{\mathbf{L}_{route}}((x^{N_s},y^{N_s}))$},
\end{equation}
where $\mathbf{1}$ is the indicator function. $\mathbf{T}$ is a trajectory represented by $N_s$ waypoints $\{(x^s,y^s,\theta^s)\}^{N_s}_{s=1}$, each indicating $x$,$y$ coordinates and heading angles. At intersections, NAVI strictly aligns with the provided navigation command. On parallel-lane roads, NAVI intentionally remains 1 for all on-route lanes, reflecting the design choice that the vehicle should understand driving intention rather than rigidly follow a single lane. In the \textbf{NavControl} dataset, NAVI is computed for every trajectory proposal under each alternative command, serving both as a simulation label for training trajectory-scoring methods and as the foundation for the controllability evaluation below.

\subsection{Controllability Measure (CM)}\label{sec:cm}
Building on NAVI, we define the Controllability Measure to evaluate a model's responsiveness to different commands at the same scene and the safety of its driving performance:
\begin{equation}
CM = \frac{1}{|C'|} \sum_{c \in C'}NAVI(\mathbf{T}^c) \cdot PDM(\mathbf{T}^c),
\end{equation}
where $C'$ is the set of all permissible driving commands for the current scenario, $\mathbf{T}^c$ is the model's output trajectory for command $c$, $NAVI(\mathbf{T}^c)$ verifies whether the trajectory's last waypoint lies on the route corresponding to command $c$, and $PDM(\mathbf{T}^c)$ is the trajectory's PDM score under command $c$. By multiplying NAVI with PDM, CM rewards only trajectories that both follow the commanded navigation direction \emph{and} drive safely—a trajectory achieving high PDM but ignoring the command receives zero credit.

\section{NaviHydra framework}\label{sec:navihydra}
As depicted in Fig.~\ref{fig:main}, NaviHydra builds on the Hydra-MDP~\cite{li2024hydra} trajectory-scoring paradigm and introduces \textbf{trajectory gathering} for context-position-aware trajectory features, while incorporating \textbf{NAVI}~(Sec.~\ref{sec:navi}) as an additional distillation head for explicit navigation compliance supervision. The framework encodes sensor data into a BEV feature $\mathbf{B}^{origin}$ via a pretrained BEV encoder, then scores $N_t$ trajectory proposals through a trajectory decoder~(Sec.~\ref{sec:td}) and a hydra scorer~(Sec.~\ref{sec:hydrascorer}).

\vspace{-2mm}
\subsection{Trajectory decoder}\label{sec:td}
The trajectory decoder constructs spatially-grounded trajectory features and enriches them with environmental context via a transformer. It comprises three key steps.

\paragraph{Trajectory gathering.}\label{sec:gathering}
Hydra-MDP~\cite{li2024hydra} uses an MLP to map trajectory proposal coordinates to query embeddings, which encodes spatial position but does not incorporate scene context along the trajectory. We augment this with BEV-based trajectory gathering: for each of the $N_t$ trajectory proposals $\{\mathbf{T}_i\}^{N_t}_{i=1}$, where $\mathbf{T}_i = \{(x_i^s,y_i^s,\theta_i^s)\}^{N_s}_{s=1}$, we first apply max pooling to $\mathbf{B}^{origin}$ to obtain a downsampled feature $\mathbf{B}^{down}$ with a larger receptive field per grid. We then gather each trajectory feature by sampling the BEV grids at the proposal's waypoint coordinates:
\vspace{-1mm}
\begin{equation}
    \mathbf{F}_i = flatten(\{\mathbf{B}^{down}(x_i^s,y_i^s)\}^{N_s}_{s=1}).
    \label{eq:inittrajfeat}
\end{equation}
By directly sampling scene features along each candidate path, trajectory gathering provides rich environmental context—such as lane boundaries, obstacles, and drivable areas—at the positions the trajectory would actually traverse, complementing the coordinate-based embeddings with perception-grounded information.

\paragraph{Navigation command integration.}\label{sec:methods:cmd}
Navigation commands $C = \{Turn~Left, Go~Straight, Turn~Right\}$ are encoded as learnable embeddings. We fuse these with grouped BEV tokens $\mathbf{B}^{grouped}$ (formed by grouping every $N_s$ neighboring grids from $\mathbf{B}^{down}$) via an SE layer~\cite{DBLP:journals/corr/abs-1709-01507}, yielding navigation-aware BEV features $\mathbf{B}^{navi} = \mathbf{SE}(\mathbf{B}^{grouped}, c)$. The command embedding is also added to each trajectory feature $\mathbf{F}_i$.
\begin{table*}[htb]
\caption{Quantitative comparison in \textbf{navtest} split of NAVSIM benchmark. $\dagger$: retrained on \textbf{navtrain}. *: evaluated using official checkpoint. $\ddagger$: retrained with our $\mathcal{V}_{4096}$ trajectory proposals. 
}
\vspace{-2mm}
\small
\centering
\begin{tabular}{l| c | c c c c c |c}

    \toprule
    Method 
    & Inputs
    & {NC$\uparrow$} 
    & {DAC$\uparrow$}
    & {EP$\uparrow$} 
    & {TTC$\uparrow$} 
    & {C$\uparrow$} 
    & {PDM Score$\uparrow$}  \\
    \midrule
    
    PDM-Closed~\cite{dauner2023parting} & Perception GT & 94.6 & 99.8 & 89.9 & 86.9 & 99.9 & 89.1    \\
    \midrule
    SSR$\dagger$~\cite{linavigation} & Camera & 93.7 & 86.3 & 70.0 & 86.5 & 98.1 & 73.6  \\
    Transfuser*~\cite{chitta2022transfuser} & LiDAR \& Camera & 97.8 & 92.1 & 78.6 & 92.8 & 100 & 83.4  \\
    UniAD ~\cite{hu2023planning} & Camera & 97.8 & 91.9 & 78.8& 92.9 &100 & 83.4   \\
    PARA-Drive ~\cite{weng2024drive} & Camera & 97.9 & 92.4 & 79.3& 93.0 &99.8 & 84.0   \\
    Hydra-MDP-$\mathcal{V}_{4096}$$\ddagger$~\cite{li2024hydra} & Camera & 98.3 & 97.3 & 84.3 & 93.4 & 100 & 89.0  \\
    Hydra-MDP++ $\mathcal{V}_{8192}$ ~\cite{li2025hydramdpadvancingendtoenddriving} & Camera & 98.6 & 98.6 & 85.7 & 95.1 & 100 & 91.0 \\
    Hydra-MDP-$\mathcal{V}_{16384}$* ~\cite{li2025generalized} & Camera & 98.5 & \textbf{98.8} & 85.4 & 94.9 & 100 & 90.8 \\
    \midrule
    NaviHydra-$\mathcal{V}_{4096}$ (Ours) & Camera & 98.4 & 97.2 & 84.0 & 95.1 & 99.5 & 89.3 \\
    NaviHydra-$\mathcal{V}_{4096}$ (Ours) & LiDAR \& Camera & \textbf{98.7}& 98.6 & \textbf{88.7} & \textbf{96.2} & 100 & \textbf{92.7} \\

    \bottomrule
\end{tabular}
\label{table:result}
\end{table*}

\begin{figure*}[t]
    \centering
        \includegraphics[width=1.0\linewidth]{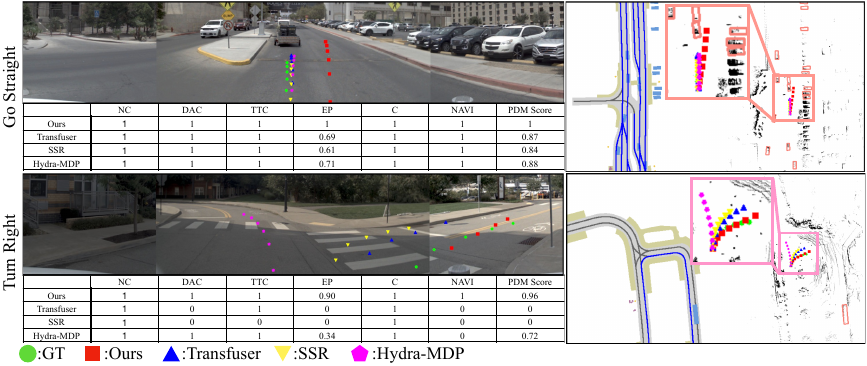}
        \vspace{-6mm}
        \caption{
        Qualitative comparison of two selected scenes in \textbf{navtest} split. Input navigation commands are listed on the left. Routes are displayed as blue lines in the semantic map. Additionally, PDM scores along with other sub-scores for the evaluated trajectories are provided. Our trajectory achieves the highest PDM-score and demonstrates the best alignment with the navigation command.
    }
    \label{fig:main_qualitative}
    \vspace{-4mm}
\end{figure*}

\paragraph{Transformer decoder.}
We flatten $\mathbf{B}^{navi}$ into environmental tokens $\mathbf{F}_{env}$ that serve as keys and values, with trajectory features $\{\mathbf{F}_i\}_{i=1}^{N_t}$ as queries to produce final trajectory features:
\begin{equation}
    \{\mathbf{F}_i^{final}\}_{i=1}^{N_t} = \mathbf{\Phi}(Q=\{\mathbf{F}_i\}_{i=1}^{N_t},\; K, V=\mathbf{F}_{env})
    \label{eq:finaltrajfeat}
\end{equation}
where $\Phi$ represents a 3-layer decoder-only transformer architecture.

\subsection{Hydra distillation}\label{sec:hydrascorer}
Following Hydra-MDP~\cite{li2024hydra}, a multi-head scorer predicts per-metric scores from the final trajectory features $\{\mathbf{F}_i^{final}\}_{i=1}^{N_t}$. Each metric $m \in M$ is handled by a dedicated MLP head, and an imitation head ($im$) is trained on L2 distance to the expert trajectory following Hydra-MDP~\cite{li2024hydra}. Crucially, we include the NAVI metric introduced in Sec.~\ref{sec:navi} as an additional distillation head, providing explicit gradient signals for navigation-compliant trajectory selection.

\paragraph{Loss terms.}
The full metric set is $M = \{NC, DAC, TTC, EP, C, NAVI\}$. The loss is formed as:
\begin{equation}
    \mathcal{L} = \sum_{m\in M\cup\{im\}} \sum_{i=1}^{N_t} k_m\mathcal{L}_{m}(\hat{\mathbf{S}}_i^m,\mathbf{S}_i^m),
\end{equation}
where we apply BCE loss for binary metrics $M_{binary} = \{NC, DAC, TTC, C, NAVI\}$ and MSE loss for $EP$.

\subsection{Inference}\label{sec:inference}
At inference time, we apply log-sigmoid to the predicted binary sub-scores and log-softmax to the imitation score before combining them. The final trajectory is selected by:
\begin{equation}
    \mathcal{S}_i^{final} = \sum_{m\in M\cup\{im\}} w_m\mathcal{S}_i^{m},
\end{equation}
where $\mathcal{S}$ is the processed score and the weighting factors $w_m$ are determined via grid search.

\section{Experiments}\label{sec:exp}
\subsection{Datasets}
\paragraph{NAVSIM.}
We utilize NAVSIM~\cite{Dauner2024NEURIPS} as our dataset for training and evaluation. The NAVSIM dataset is constructed based on the OpenScene~\cite{openscene2023} dataset, which is a redistribution of the nuPlan~\cite{nuplan} dataset. In comparison to nuPlan, NAVSIM down-samples the sensor data to 2 Hz and includes only relevant annotations, such as 2D HD maps with semantic information and 3D bounding boxes for road participants. NAVSIM provides pre-selected training and testing splits, referred as \textbf{navtrain} and \textbf{navtest}, which contain 103,288 and 12,146 samples for training/validation and testing, respectively.

\paragraph{NavControl.}
The \textbf{NavControl} dataset is constructed from the NAVSIM splits following the procedure in Sec.~\ref{sec:NavControl}. The training/validation split contains 13,406 scenarios with 30,737 direction samples, drawn from \textbf{navtrain}. The test split contains 1,512 scenarios with 3,531 direction samples, drawn from \textbf{navtest}. Each scenario is evaluated under all feasible navigation commands to assess controllability.
\vspace{-2mm}
\subsection{Metrics}\label{sec:metrics}
\paragraph{PDM score.}
The NAVSIM benchmark uses the PDM score to evaluate the closed-loop performance of the trajectory predictions, which is defined as:
\begin{equation}
    \scalebox{1.0}{$PDM = NC \times DAC \times \frac{(5\times TTC + 5\times EP +2\times C)}{12},$}
\end{equation}
where $NC, DAC, TTC, EP, C$ are the aforementioned closed-loop sub-scores. Additionally, we evaluate the model with the NAVI metric to assess if the output trajectory follows the navigation route.
\paragraph{Controllability measure.} As defined in Sec.~\ref{sec:cm}, the Controllability Measure~(CM) evaluates a model's ability to respond to different navigation commands at the same scene, rewarding both correct directional response and safety.

\vspace{-2mm}
\subsection{Baseline methods} PDM-closed~\cite{dauner2023parting} is a strong planner with privileged perception. UniAD~\cite{hu2023planning} and PARA-Drive~\cite{weng2024drive} are well-known end-to-end baselines. Hydra-MDP~\cite{li2024hydra,li2025generalized} and Hydra-MDP++~\cite{li2025hydramdpadvancingendtoenddriving} both utilize a hydra scorer to distill from the offline simulator. Transfuser~\cite{chitta2022transfuser} is a simple but effective transformer-based method. SSR~\cite{linavigation}  is a perception-task-free E2E method, utilizing an SE-layer to fuse the navigation command into the BEV feature.

\paragraph{Fair comparison protocol.}
To disentangle the gains of our NaviHydra architecture from those of the \textbf{NavControl} dataset, we design a two-level comparison. First, we retrain Hydra-MDP with our $\mathcal{V}_{4096}$ trajectory proposals using camera-only input, producing Hydra-MDP-$\mathcal{V}_{4096}$, and train the corresponding NaviHydra-$\mathcal{V}_{4096}$ under identical conditions. This isolates the architectural contribution. Second, we retrain all baselines—SSR, Transfuser, Hydra-MDP-$\mathcal{V}_{4096}$, and NaviHydra-$\mathcal{V}_{4096}$—with the additional \textbf{NavControl} intersection scenarios (denoted $\S$), isolating the data contribution. Notably, for augmented scenarios with alternative commands, we select the trajectory with the best CM out of $\mathcal{V}_{4096}$ trajectory proposals as the expert trajectory for imitation learning purposes. For Transfuser and SSR in the first group, we use the official checkpoint and our retrained model on \textbf{navtrain}, respectively.

\subsection{Implementation details}
In practice, we cluster the NAVSIM expert trajectories into $N_t=4096$ trajectory proposals with $N_s=8$ waypoints each.
Our model NaviHydra is trained on the \textbf{navtrain} split using 8 A100 GPUs with a batch size of 64 across 20 epochs. A constant learning rate of 1$\times$10$^{-4}$ is used for 20 epochs, and gradient clipping with 0.5 magnitude. For camera only model, we employ VoVNet~\cite{lee2019energy} as our image encoder and adopt a simple BEV transformer~\cite{li2025generalized} to query the BEV feature from image tokens. For LiDAR\&Camera model, we employ BEVFusion~\cite{liu2022bevfusion} as our perception backbone. The shapes of $\mathbf{B}^{origin}$ and $\mathbf{B}^{down}$ are 240$\times$160$\times$256 and 60$\times$40$\times$256 respectively. In inference time, the weighting factors for each sub-score are $w_{nc}=0.47$, $w_{dac}=0.90$, $w_{ttc}=0.99$, $w_{ep}=0.08$, $w_{c}=0.06$, $w_{navi}=0.25$, $w_{im}=0.01$.

\subsection{NAVSIM evaluation}
\paragraph{Quantitative comparison.}Results are reported in Table~\ref{table:result}. To isolate the architectural contribution of NaviHydra, we retrain Hydra-MDP with our $\mathcal{V}_{4096}$ trajectory proposals using the same camera-only setup, i.e.\ identical image encoder, transformer query dimension, and trajectory simulation labels. Under this controlled comparison, NaviHydra-$\mathcal{V}_{4096}$ (camera-only) achieves a PDM score of 89.3, surpassing Hydra-MDP-$\mathcal{V}_{4096}$ (89.0). The improvement is primarily driven by a notably higher TTC (95.1 vs.\ 93.4), indicating that trajectory gathering provides richer spatial context for collision-aware scoring, while maintaining competitive performance across all other sub-metrics. With LiDAR \& Camera fusion, our method further reaches 92.7 PDM, demonstrating the effectiveness of our trajectory decoder design.

\paragraph{Qualitative comparison.} We present a qualitative comparison of SSR \cite{linavigation}, Transfuser \cite{chitta2022transfuser}, Hydra-MDP~\cite{li2024hydra,li2025generalized} and our method in Fig.~\ref{fig:main_qualitative}. Our approach demonstrates superior performance in terms of both safety and navigation compliance indicated by the PDM score and NAVI score.
\begin{table*}[htb]
\caption{Controllability evaluation in \textbf{NavControl} test split. $\dagger$: retrained on \textbf{navtrain}. *: evaluated using official checkpoint.$\ddagger$: retrained with our $\mathcal{V}_{4096}$ trajectory proposals. $\S$: trained with additional \textbf{NavControl} intersection scenarios.}
\vspace{-2mm}
\small
\centering
\begin{tabular}{l|c|c c c c c c | c | c }

    \toprule
    Method 
    & Inputs
    & {NC$\uparrow$} 
    & {DAC$\uparrow$}
    & {EP$\uparrow$} 
    & {TTC$\uparrow$} 
    & {C$\uparrow$}
    & {NAVI$\uparrow$}
    & {PDM Score$\uparrow$}
    & {CM$\uparrow$}\\
    
    \midrule
    SSR$\dagger$~\cite{linavigation} & Camera & 91.3 & 80.6 & 67.7 & 83.3 & 95.4 & 62.8 & 68.8 & 49.9 \\
    Transfuser*~\cite{chitta2022transfuser} & LiDAR \& Camera &96.7 & 91.7 & 80.6 & 91.6 & 99.8 & 67.5 & 83.4 & 59.4 \\
    Hydra-MDP-$\mathcal{V}_{4096}$$\ddagger$~\cite{li2024hydra} & Camera& 98.0 & 97.4 & \textbf{88.4} & 95.0 & \textbf{100} & 69.6 & \textbf{91.4} & 63.8 \\
    NaviHydra-$\mathcal{V}_{4096}$ (Ours) & Camera&\textbf{98.4}& \textbf{97.8} & 87.1 & \textbf{96.0} & 99.2 & \textbf{73.9}& \textbf{91.4} & \textbf{67.6}\\
    \midrule
    SSR$\S$~\cite{linavigation} & Camera& 86.8 & 78.9 & 64.0 & 77.5 & 80.9 & 70.2 & 61.8 & 48.6\\
    Transfuser$\S$~\cite{chitta2022transfuser}& LiDAR \& Camera & 88.8 & 88.2 & 73.1 & 81.3 & 89.9 & 81.5 & 72.6 & 62.6\\
    Hydra-MDP-$\mathcal{V}_{4096}$$\ddagger$$\S$~\cite{li2024hydra} & Camera& 95.8 & 97.2 & \textbf{87.7} & 90.3 & \textbf{98.2} & 74.8 & \textbf{88.3} & 66.5 \\
    NaviHydra-$\mathcal{V}_{4096}$$\S$ (Ours) & Camera& \textbf{97.2} & \textbf{97.4} & 83.2 & \textbf{94.3} & 96.4 & \textbf{87.6} & 88.0 & \textbf{77.5} \\

    \bottomrule
\end{tabular}
\label{table:control}
\end{table*}
\begin{figure*}[t]
    \centering
        \includegraphics[width=1.0\linewidth]{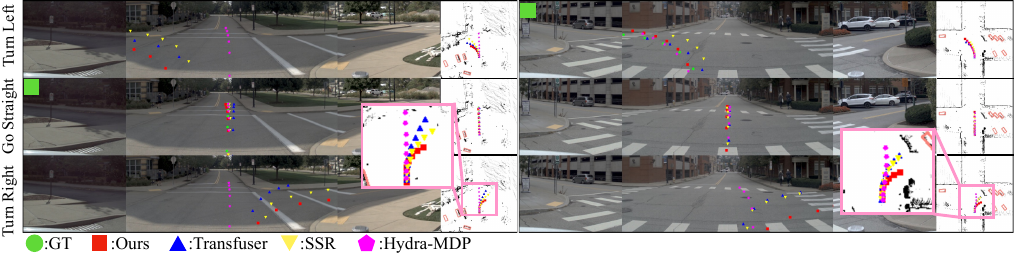}
        \vspace{-6mm}
        \caption{
        Qualitative comparison of controllability test in \textbf{NavControl} split. We present two scenarios, each of them is evaluated with 3 different navigation commands (Turn Left, Go Straight, Turn Right) as input. Only the scene corresponding to the original navigation command displays the ground truth trajectory, which is annotated with a green box.
    }
    \label{fig:control}
\end{figure*}
\begin{table*}[htb]
\caption{Ablation studies of various model components of NaviHydra in \textbf{navtest} and \textbf{NavControl} evaluation.}
\vspace{-2mm}
\small
\setlength{\tabcolsep}{1mm}
\centering
\begin{tabular}{l|cc|ccccccc|ccc}

    \toprule
    \multirow{2}{*}{ID} &
    \multicolumn{2}{c|}{Modules} &
    \multicolumn{7}{c|}{\textbf{navtest} Metrics} &
    \multicolumn{3}{c}{\textbf{NavControl} Metrics} \\

    & NAVI loss
    & Traj gathering
    & {NC$\uparrow$} 
    & {DAC$\uparrow$}
    & {EP$\uparrow$} 
    & {TTC$\uparrow$} 
    & {C$\uparrow$} 
    & {NAVI$\uparrow$}
    & {PDM Score$\uparrow$}  
    & {NAVI$\uparrow$}
    & {PDM Score$\uparrow$}
    & {CM$\uparrow$}\\
    \midrule
    1 &\xmark &\cmark & 98.4 & 98.3 & 88.2 & 95.8 & 100 & 97.8 & 92.1& 64.8 & 84.5 & 55.8 \\
    2 &\cmark &\xmark & 98.4& 98.2 & 83.9 & 95.5 & 100 & \textbf{98.9} & 89.5 & 85.9& 84.5 & 73.5\\
    3 &\cmark &\cmark & \textbf{98.7}& \textbf{98.6} & \textbf{88.7} & \textbf{96.2} & 100 & 98.0 & \textbf{92.7} & \textbf{87.6}& \textbf{88.0} & \textbf{77.5} \\

    \bottomrule
\end{tabular}
\label{table:ablation}
\end{table*}

\vspace{-1mm}
\subsection{NavControl evaluation}
We evaluate controllability on the \textbf{NavControl} benchmark, where every method receives each feasible navigation command for every intersection scenario. 

\paragraph{Quantitative evaluation.}
Tab.~\ref{table:control} reports CM and NAVI alongside PDM sub-scores. In the upper block, where all methods are trained with original \textbf{navtrain} split, Hydra-MDP scores the NAVI of 69.6, indicating more than 30\% of driving commands are ignored. NaviHydra-$\mathcal{V}_{4096}$ improves NAVI to 73.9 and achieves the highest CM of 67.6 among all four methods, outperforming Hydra-MDP-$\mathcal{V}_{4096}$ (63.8) and Transfuser (59.4), confirming the architectural advantage of trajectory gathering and NAVI supervision under the same trajectory proposals. The lower block ($\S$) reveals that augmenting training with \textbf{NavControl} intersection scenarios consistently improves NAVI across \emph{all} methods. For models with sufficient driving quality, this translates into clear CM gains—Transfuser improves from 59.4 to 62.6, Hydra-MDP from 63.8 to 66.5, and NaviHydra from 67.6 to 77.5—validating the \textbf{NavControl} dataset as a broadly effective augmentation for controllability. Notably, NaviHydra$\S$ benefits the most (+9.9 CM), demonstrating that our framework is specifically designed to exploit the contrastive intersection-level signal.
SSR~\cite{linavigation}, which lacks auxiliary perception tasks, achieves the lowest PDM score among all methods, reflecting its weaker closed-loop driving performance. Nevertheless, SSR still attains a competitive NAVI score of 62.8 thanks to its SE-layer-based navigation command fusion, and improves further to 70.2 after \textbf{NavControl} augmentation.

We observe that PDM scores on the \textbf{NavControl} test split decrease after intersection-data augmentation for all methods (e.g.,\ Hydra-MDP: 91.4$\to$88.3; NaviHydra: 91.4$\to$88.0). We attribute this to the increased difficulty of intersection scenarios: turning maneuvers require precise lane changes and tighter spatial margins, making collisions, drivable-area violations, and comfort degradation more likely than in straight-driving segments. The additional intersection data therefore exposes the planner to harder cases that lower the average PDM. However, this trade-off is deliberate—the substantial NAVI improvements (Hydra-MDP: 69.6$\to$74.8; NaviHydra: 73.9$\to$87.6) and the corresponding CM gains demonstrate that models trained with intersection data produce trajectories that actually follow the commanded direction, which is the primary goal of a controllable planner.
\paragraph{Qualitative evaluation.}
Depicted in Fig.~\ref{fig:control}, we provide visualizations of 2 selected samples from \textbf{NavControl}. Our method demonstrates high driving safety and controllability, benefiting from the NAVI-based navigation compliance supervision. Hydra-MDP, without using navigation compliance as a supervision signal, fails to respond correctly to the input navigation commands. Transfuser and SSR, while following the navigation commands, fall short of drivable area compliance.
\vspace{-2mm}
\subsection{Ablation studies}
We conduct several ablation studies in Tab.~\ref{table:ablation} using PDM-evaluation in \textbf{navtest} split and using controllability test in the \textbf{NavControl} test split to validate the effectiveness of the following building blocks. We use LiDAR \& Camera model for \textbf{navtest} and camera-only model for \textbf{NavControl} as baselines.

\paragraph{NAVI loss.} Shown in Tab.~\ref{table:ablation} between ID 1 and 3, incorporating the NAVI loss introduced in Sec.~\ref{sec:hydrascorer}, increases the PDM score in \textbf{navtest} by 0.6, and CM in the \textbf{NavControl} test split by 21.7. The CM and PDM-Score are overall improved after incorporating the NAVI loss, demonstrating the positive impact of this building block facing alternative navigation commands.

\paragraph{Trajectory gathering.}Illustrated by ID 2 and 3 in Tab.~\ref{table:ablation}, we observe a significant improvement in both PDM-score and the controllability measure after incorporating the trajectory gathering method introduced in Sec.~\ref{sec:td}, confirming that trajectory gathering improves trajectory feature quality and downstream planning performance.

\section{Conclusion}
In this paper, we identify a critical navigation compliance gap in current trajectory-scoring planners, caused by the absence of intersection-level contrastive data and a definitive route-adherence metric. To close it, we construct the \textbf{NavControl} dataset of nearly 15\,k intersection scenarios augmented with alternative commands and routing annotations, introduce the binary \textbf{NAVI} metric and the derived \textbf{Controllability Measure (CM)}, and propose the \textbf{NaviHydra} framework whose \textbf{trajectory gathering} and NAVI-supervised hydra distillation exploit this contrastive signal. Fair-comparison experiments controlling for trajectory proposal count and training data confirm that both the \textbf{NavControl} dataset and the NaviHydra architecture contribute independently to improved controllability and safety, achieving state-of-the-art results on both the NAVSIM and the \textbf{NavControl} benchmarks.
Future research will investigate more diverse planners including Diffusion-based planners and VLM-assisted planners under reactive simulation for navigation compliance. 

\end{document}